\documentclass[a4paper, 10pt, conference]{ieeeconf}      % Use this line for a4
\usepackage{FG2024}
\usepackage{graphicx,balance}

\FGfinalcopy % *** Uncomment this line for the final submission

\IEEEoverridecommandlockouts                              % This command is only
\usepackage{amsmath} % assumes amsmath package installed
\usepackage{amsfonts}
\usepackage{multirow}
\usepackage{float}
\usepackage{hyperref}
\usepackage{xcolor}
\usepackage{siunitx}

\usepackage{bbm}

\def\FGPaperID{64} % *** Enter the FG2024 Paper ID here

\title{\LARGE \bf
Subject-Based Domain Adaptation for Facial Expression Recognition
}

\author{\parbox{16cm}{\centering
    {\large Muhammad Osama Zeeshan$^1$, Muhammad Haseeb Aslam$^1$, Soufiane Belharbi$^1$, Alessandro Lameiras Koerich$^1$, Marco Pedersoli$^1$, Simon Bacon$^2$, Eric Granger$^1$}\\
    {\normalsize
    $^1$  LIVIA, Dept. of Systems Engineering, ETS Montreal, Canada\\
    $^2$ Dept. of Health, Kinesiology \& Applied Physiology, Concordia University, Montreal, Canada}
    \small{muhammad-osama.zeeshan.1@ens.etsmtl.ca,  eric.granger@etsmtl.ca}
    }
}

\usepackage{fancyhdr}

\begin{document}

\ifFGfinal
\thispagestyle{empty}
\pagestyle{empty}
\else
\author{Anonymous FG2024 submission\\ Paper ID \FGPaperID \\}
\pagestyle{plain}
\fi
\maketitle

\makeatletter
\newcommand{\thickhline}{%
    \noalign {\ifnum 0=`}\fi \hrule height 1pt
    \futurelet \reserved@a \@xhline
}
\newcolumntype{"}{@{\hskip\tabcolsep\vrule width 1pt\hskip\tabcolsep}}
\makeatother
\thispagestyle{fancy}

%%%%%%%%%%%%%%%%%%%%%%%%%%%%%%%%%%%%%%%%%%%%%%%%%%%%%%%%%%%%%%%%%%%%%%%%%%%%%%%%
\begin{abstract}
%% Application and key challenges
Adapting a deep learning model to a specific target individual is a challenging facial expression recognition (FER) task that may be achieved using unsupervised domain adaptation (UDA) methods. Although several UDA methods have been proposed to adapt deep FER models across source and target data sets, multiple subject-specific source domains are needed to accurately represent the intra- and inter-person variability in subject-based adaption. This paper considers the setting where domains correspond to individuals, not entire datasets. Unlike UDA, multi-source domain adaptation (MSDA) methods can leverage multiple source datasets to improve the accuracy and robustness of the target model. However, previous methods for MSDA adapt image classification models across datasets and do not scale well to a more significant number of source domains.  
%In recent years, there has been a growing demand for the adoption of facial expression recognition (FER) across many domains. Nonetheless, the challenge lies in the significant variability of facial expressions among individuals due to cultural, ethnicity, or capturing conditions. This leads to a substantial disparity between data used for training the model and that used for testing. The conventional approach of adapting or fine-tuning a model using fully labeled data requires annotating each sample, a process that is costly and unfeasible. Unsupervised domain adaptation (UDA) has emerged as a promising approach for leveraging unlabeled data in FER. However, current single UDA techniques have only considered a single source that adapts to the target, which limits their capacity to handle target variations and diversification. To address this challenge, multi-source domain adaptation (MSDA) has gained significant popularity and incorporates information from multiple sources to enhance models' resilience to different target variations. 
%%
%% Proposal
This paper introduces a new MSDA method for subject-based domain adaptation in FER. It efficiently leverages information from multiple source subjects (labeled source domain data) to adapt a deep FER model to a single target individual (unlabeled target domain data). 
%In this paper, a new approach is introduced to address the challenge of unlabeled data in FER by incorporating information from multiple source subjects. Specifically, we proposed a Multi-Subject Domain Adaptation (MS\textsuperscript{b}DA) technique that leverages information from multiple subjects (as labeled source domains) and then adapts to a single individual (as an unlabeled target domain). 
%This paper also integrates a new strategy to generate augmented confident pseudo-labels for the unlabeled target subject. 
During adaptation, our subject-based MSDA first computes a between-source discrepancy loss to mitigate the domain shift among data from several source subjects. Then, a new strategy is employed to generate augmented confident pseudo-labels for the target subject, allowing a reduction in the domain shift between source and target subjects. 
%%
%% Experimental result
%To evaluate the efficacy of our proposed methodology, we conducted a comprehensive set of experiments on the BioVid heat and pain dataset (PartA), which comprises 87 subjects, treating them as labeled sources and unlabeled target domains.
% Experiments\footnote{\textcolor{red}{\textbf{Supplementary material} contains our code, which will be made public, and additional experimental results.}} on the challenging 
% BioVid heat and pain dataset (PartA) with 87 subjects shows that our Subject-based MSDA can outperform state-of-the-art methods yet scale well to multiple subject-based source domains. 
Experiments\footnote[1]{Our code: \url{https://github.com/osamazeeshan/Subject-Based-Domain-Adaptation-for-FER}}
performed on the challenging BioVid heat and pain dataset with 87 subjects and the UNBC-McMaster shoulder pain dataset with 25 subjects show that our subject-based MSDA can outperform state-of-the-art methods yet scale well to multiple subject-based source domains.
\end{abstract}

%%%%%%%%%%%%%%%%%%%%%%%%%%%%%%%%%%%%%https://www.overleaf.com/project/64ee281c61c9bf6f750cc8b9%%%%%%%%%%%%%%%%%%%%%%%%%%%%%%%%%%%%%%%%%%%

% \begin{figure*}[t!]
% \centering
% \includegraphics[width=0.88\textwidth]{combined_vs_multi.png}
% %\caption{\textbf{Model proposed for A-V fusion based on joint cross-attention based . "Inc" in V feature extractor denotes Inception Module.}}
% \caption{\textbf{Proposed network for \textit{Multi-Subject Unsupervised Domain Adaptation (MS\textsuperscript{b}UDA).}}}
% \label{fig:combined_vs_multi}
% \end{figure*}

% \begin{figure}[h]
%     \centering
%     \includegraphics[width=0.88\textwidth]{stage_1_mmd.png}
%     \caption{Stage-1, Aligning sources and target using adaptation loss with a supervised loss from source domains}
%     \label{fig:stage_1_mmd}
% \end{figure}

%MP: to me the intro should be:
% -1 Normally domain adaptation is used for different domains such as devices, working conditions etc..
% -2 For FER and in general all tasks that deal with individuals, the largest source of shift is the individual (show it with a t-SNE plot).
% -3 In this work we show that by just using UDA for individuals, in which the target domain is an individual, we can already and clearly improve the performance of any classification method for FER.
% -4 Furthermore, as for training we deal with multiple IDs, a multi-source approach can further boost performance as it aligns all source data.
% -5 However, common MS-UDA methods cannot scale well with so many domains (in the order of hundreds or thousands), so we propose something that deals with that...

\section{Introduction}
% Application intro
%Given the advancement in deep learning (DL) over the past decade, many applications sought to develop reliable, sustainable, and intelligent systems \cite{hu2018squeeze, simonyan2014very}. 
Facial expression recognition (FER) techniques have gained popularity in several applications, such as pain estimation in health care, detecting suspicious or criminal behavior, and sleep or stress estimation \cite{han2020personalized, li2018deep}. In e-health, automatic detection of facial activity linked to, e.g., pain is essential for people with communication problems, those who suffer from severe illnesses, or those who have brain injuries \cite{praveen2020deep, rajasekhar2021deep}.  %Therefore, the need for a FER system is growing substantially.

%--- Problem with conventional DL method and Introduce UDA 
Deep learning (DL) models provide state-of-the-art performance in many computer vision tasks, such as object detection \cite{ren2015faster}, image classification \cite{hu2018squeeze, simonyan2014very}, and semantic segmentation \cite{long2015fully}. However, these tasks typically rely on large-scale datasets and supervised learning to train an accurate model. Moreover, the model is assumed to be trained and tested on data sampled from the same underlying distribution. When a model is applied to test data that diverges from the training set, its performance can decline significantly.  
%This phenomenon is called domain shift \cite{ben2010theory}. 
Unsupervised domain adaptation (UDA) \cite{gretton2012kernel, long2016unsupervised, tzeng2014deep, xu2018deep} techniques have been developed to address the domain shift problem \cite{ben2010theory} between training (source) and testing (target) domain data.  
Given the cost of annotating data collected from an operational target domain, researchers have proposed many UDA methods to find a common representation space between data distributions of a labeled source domain and an unlabeled target domain. 
% The UDA techniques are based on learning the domain discriminative features on the target (test) domain to have a more robust model.
Facial expressions vary significantly depending on the individual (e.g., their expressiveness) and capture conditions, which decreases the performance of FER systems. Recently, UDA models have gained significant popularity in FER tasks \cite{han2020personalized, li2018deep, zhu2016discriminative} to detect emotions such as happiness or sadness to deal with the problem of domain-gap among distinct distributions. 
\begin{figure*}[t!]
\centering
\includegraphics[width=0.88\linewidth]{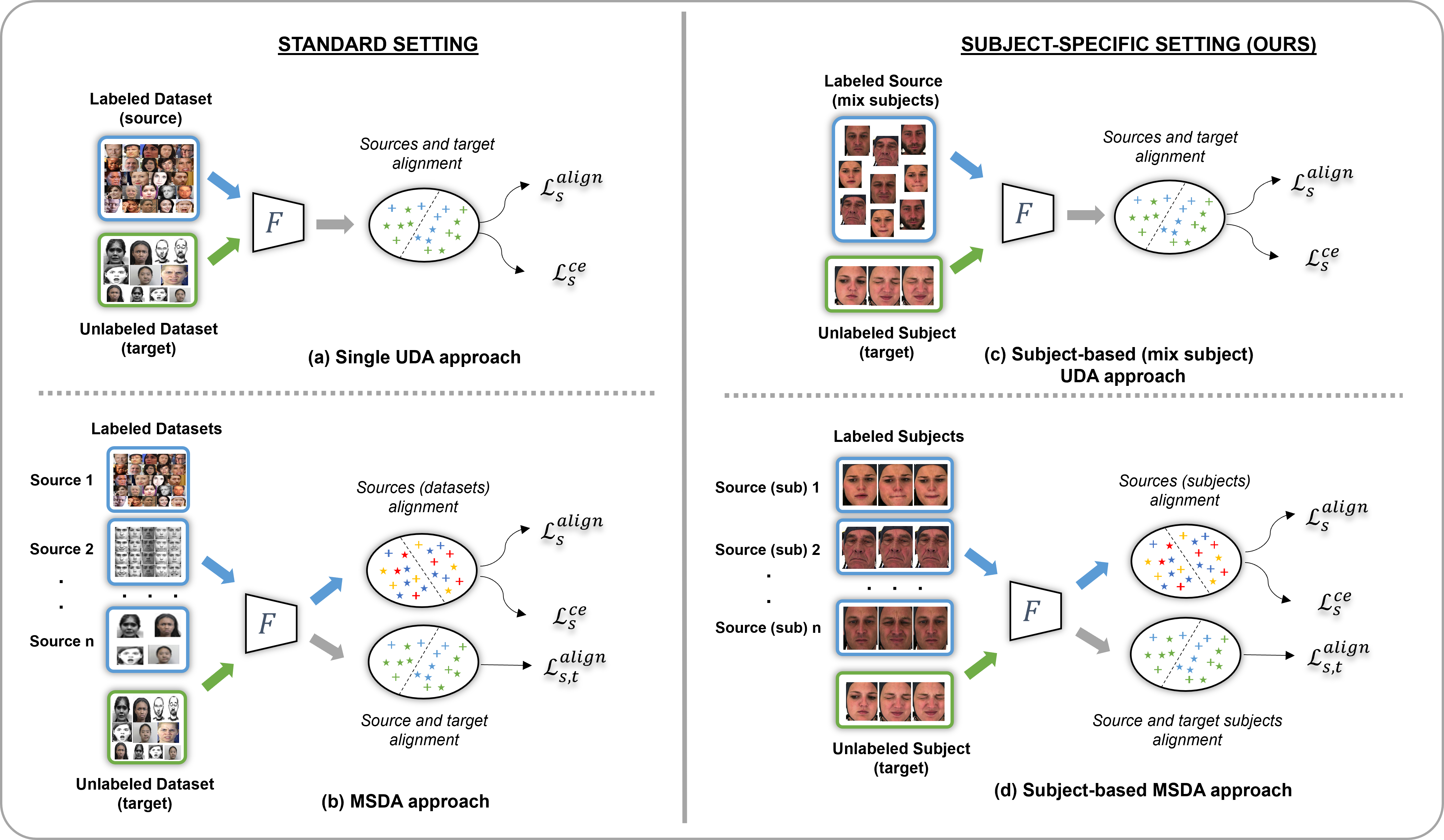}
\caption{The settings for domain adaptation of a deep FER model. On the left side, we have the standard setting that includes two approaches \textbf{(a)} single UDA approach, where one labeled dataset is adapted to a single unlabeled dataset. \textbf{(b)} The MSDA approach aligns multiple source datasets and adapts to the single target domain. On the right side, we have the subject-specific settings with two approaches. \textbf{(c)} The mix-subject UDA approach considers a single labeled source with different subject identities (IDs) aligned with unlabeled target subjects. \textbf{(d)} Subject-based multi-source domain adaptation considers each subject as a separate domain, mitigating the domain shift among the sources, and aligns data from selected sources with the target subject. \textbf{Blue} indicates labeled source data, \textbf{green} color indicates unlabeled target data, and \textbf{grey} color indicates data from both domains.}
\label{fig:trad_uda_msda}
% ALEKOE
\vspace{-10pt}
\end{figure*}

% Probelm with UDA for FER and Introduce MSDA
UDA methods focus on a single source to adapt to the target domain. However, in a more realistic scenario, data can come from various sources with varying distributions. One way is to combine all the sources to create a single domain space and adapt to the target domain, but that does not consider the domain shift between the sources that led to the decline in models' performance. 
% Recent research in psychology \cite{barrett2007experience, kitayama2006cultural, nisbett2003culture} and neuroscience shows that facial expressions are culture-dependent and not versatile, such as in pain recognition where each person is coming from a variety of distributions having different variability in expressing pain.
MSDA methods \cite{xu2018deep, deng2022robust, ren2022multi, kang2020contrastive} allow for the creation of a robust model by integrating information from a variety of source domains that can adapt to an unlabeled target domain. However, in FER task defining a dataset per domain such as RAF-DB \cite{li2017reliable}, AffWild \cite{kollias2019deep}, SEWA \cite{kossaifi2019sewa}, FER2013 \cite{goodfellow2013challenges}, or AffectNet \cite{mollahosseini2017affectnet} in a UDA or MSDA setting as shown in Fig.~\ref{fig:trad_uda_msda} (left-side) does not consider each subject variability which is important in developing a subject-specific model. These FER datasets comprise uncontrolled, in-the-wild data that lack subject-specific information. Instead, they confine diverse samples from random individuals, offering valuable data variety but not emphasizing creating a model that generalizes well to a single individual. This limitation arises from the absence of subject-based variations within the database, a crucial aspect of developing subject-specific models. 

Research in psychology and neuroscience \cite{barrett2007experience, kitayama2006cultural, nisbett2003culture} have shown that facial expressions can vary considerably from person to person.  
%are person-dependent and not versatile. 
For instance, in pain estimation applications, each person has a distinct way of expressing their pain. When dealing with subject-based data, it is therefore important to learn subject-specific representations of expressions to model the intra- and inter-person variability. Moreover, the current MSDA methods are trained on generic classification benchmark datasets \cite{li2017deeper, saenko2010adapting}, where domains correspond to entire datasets, and the number of domains is limited. This limits their ability to scale to a large number of source domains. 

% Furthermore, in FER, defining a dataset per domain such as RAF-DA, FER2013, or AffectNet in a UDA or MSDA setting (as shown in Fig. \ref{fig:trad_uda_msda}) that does not consider information about each subject consisting of different people is not feasible. Recent research in psychology \cite{barrett2007experience, kitayama2006cultural, nisbett2003culture} and neuroscience shows that facial expressions are culture-dependent and not versatile, such as in pain recognition where each person is coming from a variety of distributions having different variability in expressing pain. It is important to develop such a model that can adapt to a person that represents intra- and inter-person variability. 

This paper focuses on leveraging a larger number of source domains, where each domain consists of a distinct subject rather than datasets. We proposed an MSDA method for \emph{Subject-Based Domain Adaptation} that can adapt a deep FER model to an individual by defining each subject as a distinct domain -- labeled data from multiple subjects (source domains) and unlabeled data from a target individual (target domain). The subject-specific setting is illustrated in Fig.~\ref{fig:trad_uda_msda} (right). The subject-based UDA approach (Fig.~\ref{fig:trad_uda_msda}c) relies on a single source and target, where the source domain comprises different persons, and the model is adapted to a single target subject. In the subject-based MSDA approach (Fig.~\ref{fig:trad_uda_msda}d), each subject is linked to a distinct domain, thus multiple sources and a single target subject. The target domain consists of samples that belong to a single person, where the data is captured in a stationary environment and incorporates less effective diversity. 

In this paper, we propose an MSDA method for subject-based domain adaptation that allows including distinct subjects as source domain data to explicitly leverage their diversity, and thereby provide better adaptation to the target subject. 
%Although these multiple subjects add diversity in training a more generalized model, it raises a domain-shift problem between them. This required reducing the discrepancy among various subject distributions. 
Aligning subject class distributions between source and target domains bridges the gap between diverse individuals. However, aligning the class representation of multiple individuals to a common domain-agnostic feature space raises several challenges during MSDA, especially when the target is unlabeled. Therefore, generating the target pseudo-labels (PLs) is an important step in the alignment process. 
% Since the target domain data is unlabeled, the target and source class distribution alignment may be challenging. Therefore, generating the target pseudo-labels (PLs) is an essential component in the alignment process. 
The most common approach in MSDA for image classification \cite{scalbert2021multi, venkat2020your, yuan2022self, zhao2020multi} predicts a target PL from a source classifier combined with a confidence threshold. Still, this approach often results in noisy labels that impact the domain alignment procedure. To address this issue, we introduce the Augmented Confident Pseudo Label (ACPL) strategy that generates PLs based on augmented thresholding, producing highly reliable target PLs before training the target adaption model. 
Our contributions are summarized as follows. (1) A novel \emph{Subject-based Domain Adaptation} method is introduced to leverage the diversity of data from multiple subjects (each one considered as a source domain) for adapting to an unlabeled target subject. (2) An ACPL strategy is proposed to generate highly reliable pseudo-labels before the target adaptation without adding complexity to the training procedure. (3) Our experimental results and ablations show that the proposed subject-based domain adaptation method can outperform source-only and state-of-the-art MSDA methods on the Biovid and UNBC-McMaster datasets and scales well to a large number of source domains. Results also indicate that selecting the relevant source subjects (domains) drastically improves the model performance on an unlabeled target subject. 

\section{Related Work}

%%%
\subsection{Unsupervised Domain Adaptation}

Adapting visual DL models to operational target domains is a challenging task, commonly achieved using UDA methods, where labeled data from a single source and unlabeled data from a single target are available for adaptation. Several discrepancy-based \cite{zhu2016discriminative,han2020personalized}, reconstruction-based \cite{ghifary2016deep, zhuang2015supervised, zhu2017unpaired}, and adversarial-based \cite{chen2021cross, yan2018unsupervised} UDA methods have been proposed in the literature.

\noindent \textbf{Discrepancy-based:} Zhu et al.~\cite{zhu2016discriminative} seek to minimize the discriminative feature among the domains to diminish the domain shift. The disparity between the means of domains is determined to analyze the difference in both distributions. In \cite{han2020personalized}, a model is personalized for FER using an incremental broad learning system. Maximum mean discrepancy (MMD) and entropy are combined to measure the difference between source and target features. 

\noindent \textbf{Reconstruction-based:} In \cite{ghifary2016deep} authors presented a deep reconstruction classification model, which encodes the representation of source and target domains, labeled source data classification via supervised learning, and unlabeled target data reconstruction through unsupervised learning. Zhuang et al.~\cite{zhuang2015supervised} presented a transfer learning-based autoencoder technique containing the embedding and label encoding layers. KL divergence minimizes the distribution distance among the domains in the embedding layer. In contrast, the source domain is calculated in a supervised way. In \cite{zhu2017unpaired}, the author proposed a dual learning approach that transfers a source representation's learning into a target representation. It uses dual generators that learn the image translation mapping with two discriminators, which measure the authenticity of a newly rendered image using an adversarial loss.

\noindent \textbf{Adversarial-based:} Barros et al.~\cite{chen2021cross} proposed an adversarial graph representation adaptation approach containing two convolutional graphs stacked together to train local and domain-invariant features across domains. In \cite{bozorgtabar2020exprada}, the proposed method generates target data using an augmentation technique. AC-GAN \cite{odena2017conditional} dealt with the domain shift between newly generated images and the source domain. A dictionary learning approach is proposed in \cite{yan2018unsupervised}. 

%Critical analysis:
Although several UDA methods have been proposed to adapt FER models across datasets, several source domains are desirable in subject-based adaption to represent the intra- and inter-person diversity. In literature \cite{deng2022robust, peng2019moment, ren2022multi}, it was found that incorporating multiple source domains would help in the target adaptation process, as the samples collected from diverse sources exhibit diverse affective information.
% These approaches are widely adapted to recognize expressions, mainly concise of single source and target domain.
% simple emotions like smiling, happiness, or sadness are not applied in the classification of pain.

%%%%
\subsection{Multi-Source Domain Adaptation}

 %In UDA, the model only learns from a single labeled source to adapt to the unlabeled target domain. 
 Unlike UDA, MSDA leverages multiple source datasets to improve the accuracy and robustness of the target model. MSDA models learn from multiple labeled sources to have better generalization on the target domain. Many MSDA methods have been proposed for image classification. Peng et al.~\cite{peng2019moment} present a novel approach for adapting a model trained on a multi-source domain to perform well on the target domain. The proposed moment matching for multi-source domain adaptation (M\textsuperscript{3}SDA) technique aligns the first and second-order statistics of the feature representations of the source and target data as the objective function for domain adaptation.
Similarly, in \cite{kang2020contrastive}, the author proposed the contrastive adaptation network method, which minimizes contrastive domain discrepancy across domains. It calculates the MMD between domain distributions. K-means clustering was used to assign a PL to the target samples.
 Adversarial-based methods have also gained popularity in MSDA. Nguyen et al.~\cite{nguyen2021stem} proposed an adversarial learning student-teacher model. The network comprises a source teacher and a target student model. The ensemble-based approach is applied to achieve the teacher model by taking an average of source models. The generator is trained on the target sample, and the discriminator is with source samples that try to decrease the gap between the target and source distributions. Another similar technique was introduced by Zhao et al.~\cite{zhao2021madan} proposed a multi-source adversarial domain aggregation network, which focused on aligning the different sources and a target domain via pixel-level alignment. 

 Although these methods perform well on classification tasks, they add complexity to the model by adding additional auxiliary losses, which require careful hyperparameter tuning and increase the training time. Venkat et al. \cite{venkat2020your} present a self-supervised-based approach that does not require additional auxiliary loss to align sources and target domains. The model was first introduced to multiple sources to train a classifier for each source, then presented a target domain to the network to assign a PL based on the source classifiers' agreement rate. In \cite{deng2022robust}, authors propose the bi-level optimization (BORT\textsuperscript{2}) method, which does not require training a source domain but instead utilizes any existing single UDA network. BORT\textsuperscript{2} uses that model to generate a target PL while eliminating noisy samples that train a robust target model. Scalbert et al.~\cite{scalbert2021multi} presented contrastive learning to generate weak and strong augmented images of the target domain. It generates target labels, a hard PL is applied for the prediction from the weak augmented image, and a soft label is applied to the strongly augmented image. A cross-entropy loss is calculated among the samples whose probability is above a certain threshold.

%Critical analysis:   
% - Other MSDA methods are for general classification problems and do not scale well to several source domains.  
Despite the performance of MSDA methods on benchmark classification tasks, few methods have been proposed for FER. Unlike the general MSDA methods that only deal with a small number of source domains (up to 6), our approach is developed to scale well with the increasing number of domains. For the generation process, we introduce the ACPL technique that improves the reliability of target samples.

 % These MSDA methods perform well on image classification tasks tested on benchmark datasets. However, MSDA techniques were not yet explored in a FER setting. Thus, we are proposing a new technique for FER where leveraging multiple subjects as domains to adapt to the unlabeled target subject. These domains in our proposed setting are in large numbers (up to 87) as compared to existing MSDA techniques, where the domains were limited (up to six).
 % The current method relies on training multiple classifiers for each source to adapt, which is computationally expensive when having large domains. On the other hand, generating a target pseudo-label using a source classifier will produce noisy labels that were used to train the model and hope to eliminate during training. Our proposed approach is based on a two-step training protocol as presented in the BORT\textsuperscript{2} \cite{deng2022robust}, but we perform label denoising before the target adaptation. This way, we only train the model with the most reliable target samples and reduce the domain-shift between subject domains during training without worrying about the noisy samples.

\addtolength{\textheight}{-2cm}

%%%%%%%%%%%%%%%%%%%%%%%%%%%%%%%%%%%%%%%%%%%%%%%%%%%%%%%%%%%%%%%%%%%%%%%%%%%%%%%%
\section{Proposed Methodology}
% Discrepancy-based methods \cite{peng2019moment, kang2020contrastive, xu2021multi} have shown tremendous success in the MSDA setting. Inspired by this approach, we also proposed our method based on maximum mean discrepancy (MMD).
% has been used as a baseline for many multi-source domain adaptation techniques. Inspired by discrepancy-based methods, we also define our baseline using the maximum mean discrepancy (MMD) approach

\begin{figure*}[t!]
\centering
\includegraphics[width=0.63\linewidth]{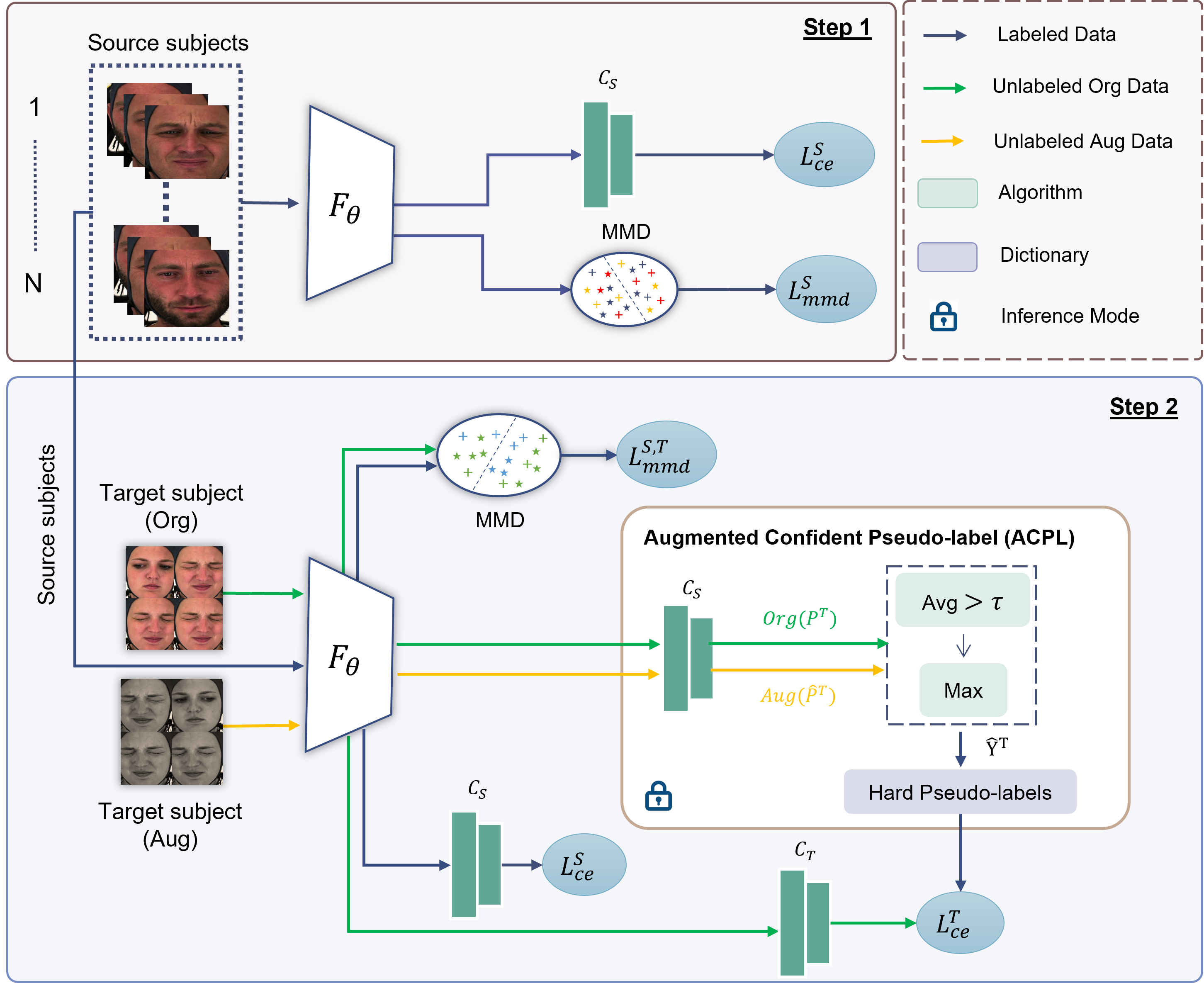}
\caption{An illustration of our proposed subject-based MSDA method. In the first step, we align labeled source subjects using discrepancy and supervision losses. In the second step, the augmented confident pseudo-label (ACPL) strategy is applied to generate reliable target PLs and, finally, train the adaptation model using the source subjects and reliable target samples.}
\label{fig:MSubDA_network}      
% ALEKOE
\vspace{-10pt}
\end{figure*}

%This section presents our MSDA method for \emph{Subject-Based Domain Adaptation}. 
The overall pipeline of the proposed MSDA method for subject-based domain adaptation is illustrated in Fig.~\ref{fig:MSubDA_network}. The MSDA two-step training strategy \cite{deng2022robust, venkat2020your} has shown tremendous success in the adaptation process. Inspired by that, we follow the same approach in aligning source and target subjects. In the first step, we align different subjects that are considered as a domain using a discrepancy loss that reduces the domain shift between each individual. A supervised loss is also calculated for every subject, and the model is jointly optimized with both losses. In the next step, we want our model to adapt to the unlabeled target subject. However, we must train it with reliable samples only to have better adaptability. Therefore, we propose an ACPL strategy to minimize the ratio of noisy labels while selecting only the most confident PLs. In ACPL, we create two image versions of each target sample by performing augmentation on every sample, denoted as \textit{Original} and \textit{Augmented} versions. We predict for every set of target samples using a trained source classifier that generates output logits, which were converted into probabilities with a \emph{softmax} layer. Finally, we compute the average probability of the two images and apply a certain threshold ($\tau$). The sample is considered reliable if this average probability exceeds the $\tau$. We then assign a PL to each target sample based on the version (either original or augmented) with the higher probability.

% We select confident target labels in inference mode by loading the source subject pre-trained model where we introduce both the original and augmented version of each sample to the train source classifier, generating output logits that are then converted into probabilities with a \emph{softmax} layer.}

% Finally, we take the average of two probabilities (original and augmented version of each sample) and apply a threshold ($\tau$) to the average of these probabilities. If the maximum class probability is greater than $\tau$, then we select the label of the sample with the highest probability as a subject pseudo-label. 

After selecting the confident samples, the adapted model is jointly trained using target and source subjects. The network is optimized using cross-entropy and discrepancy losses between source and target subjects, calculated using the MMD \cite{sejdinovic2013equivalence}.

\subsection{Notation}
Let $D^{S}=\left \{ D^{S}_1, ..., D^{S}_s \right \}$ be a multi-labeled source subject domains, and $D^{T}$ be an unlabeled target subject domain. We define ${{X_{i}^{S}}}=\left \{ x^{S}_{i_1}, ..., x^{S}_{i_n} \right \}$, where $i = \left \{1, ..., s\right \}$ the number of source domains and $n$ is the number of samples in each domain. ${{X_{n}^{T}}}=\left \{ x^T_{1}, ..., x^T_{n} \right \}$ is the input space of samples from $D^{T}$. For $D^{S}$ we also define ${{Y_{i}^{S}}}=\left \{y^{S}_{i_1}, ..., y^{S}_{i_n} \right \}$ as the output label space. We also define a feature extractor $F_\theta$ that is common in both source alignment and target adaptation. $C_S(.)$ is a source classifier that is common in all subject domains, and $C_T(.)$ is defined as a target classifier. 

% To generate the augmented version of each sample in the target domain, $X^T_n$ and $\widehat{X}^T_n$ are defined as an original and augmented version, respectively.
% We consider $n_{s}$ multiple labeled source subjects (domains) $[S_{i}]_{i=1}^{n_{s}}$ with image and label pairs $(x_{s}^{i},y_{s}^{i})_{i=1}^{n_{s}}$ and one unlabeled target subject (domain) $\mathit{T}$ consisting of only images $(x_{t}^{i})_{i=1}^{\textit{T}}$. The model is then jointly trained on $\textrm{M}_\Psi  =S_{i}\cup T$ and evaluated on the target test dataset. 

%%%%%%%%%%
\subsection{Alignment of Source Domains}

In the first step, labeled data from the source subjects is considered. The model learns using supervised loss from sources while calculating the MMD \cite{sejdinovic2013equivalence} to reduce the disparity among subjects. To reduce the domain shift between subjects, we aim to decrease the discrepancy between subject representations. The MMD is estimated in a reproducing kernel Hilbert space, defined as:

% {\small
% \begin{equation} \label{eq:sources_mmd} 
% MMD(N_{}^{S}, 
% N^{\widetilde{S}})= \mathrm{\left\| \mathbb{E}_{X^{S}}^{} [F_\theta(\mathrm{X}^{S})]-\mathbb{E}_{X^{\widetilde{S}}}^{} [F_\theta(\mathrm{X}^{\widetilde{S}})] \right\|}_{\mathcal{H}}^{2}
% \end{equation}}
{\small
\begin{equation} \label{eq:main_mmd}
\begin{split}
    MMD(P(A), 
Z(B)) & = \sum_{u,v}^{n} \frac{k(a_u,a_v)}{n^2} \\ 
& + \sum_{u,v}^{m} \frac{k(b_u,b_v)}{m^2} - 2\sum_{u,v}^{n,m} \frac{k(a_u,b_v)}{nm} 
\end{split}
\end{equation}
}
% \end{multline*}

% followed by applying confidence thresholding to distribute the target examples into confident $\mathrm{T_{cl}}$ and weak pseudo-labels $\mathrm{T_{wl}}$. 
% \begin{figure}[h]
%     \centering
%     \includegraphics[width=0.88\textwidth]{stage_1_mmd.png}
%     \caption{Stage-1, Aligning sources and target using adaptation loss with a supervised loss from source domains}
%     \label{fig:stage_1_mmd}
% \end{figure}
% We reduce the discrepancy between multiple source domains coming from $\mathrm{n_{s}}$, and it was defined as:
% \begin{equation} \label{eq:mmd} 
% \mathcal{L}_{mmd}^{s}=\mathrm{\left\| \mathrm{{E}}_{\mathrm{\mathrm{x}_{i}{}\sim n_{s}}_{}^{}}^{} [{f}(\mathrm{x}_{i}^{c})]-\mathrm{{E}}_{\mathrm{x}_{j}\sim n_{s}}^{} [{f}(\mathrm{x}_{j}^{c})] \right\|}_{\mathcal{H}}^{2} .{1}(i\neq j)
% \end{equation}
\noindent where $P(A)$ and $Z(B)$ are the features from two different subject domains $A$ and $B$, and $a_u$ and $b_u$ are the $u^{th}$ sample feature vector extracted with $F_{\theta}$ from $A$ and $B$. $k(. , .)$ denotes kernel (such as Gaussian). $n$ and $m$ represent the total number of data samples in $P(A)$ and $Z(B)$. To align the source subjects, the MMD loss is estimated between the two different subject feature vectors. We denote $F^S_A = F_{\theta}(X^S_A)$ and $F^S_B= F_{\theta}(X^S_B)$. The MMD estimation is defined as:
\begin{equation} \label{eq:src_mmd}
    L_{\mbox{mmd}}^{S} = {MMD}(F_{A}^{S}, F^S_B)
\end{equation}
\noindent where $F^S_A$ and $F^S_B$ are the extracted features from two distinct source subject domains $D^{S}_{A}$ and $D^S_B$. Minimizing the discrepancy loss aligns dissimilar subjects closer to one another in a feature space. We also minimize the categorical cross-entropy loss by using the label information:
 % whereas $\mathrm{X^{S}}$ and $\mathrm{X^{\widetilde{S}}}$, represents different samples from them. $F_\theta(.)$ is the feature vectors corresponding to different subject distributions that are required to calculate the MMD distance between them.
% \begin{equation} \label{eq:lce} \mathrm{\mathcal{L}}_{ce}^{s} = -\frac{1}{\mathrm{N}^{S}} \sum_{i=1}^{\mathrm{N}^{S}}\mathrm{y}_{i}^{}log 
% [\sigma(F(x_{i}^{s}))]
% \end{equation}
\begin{equation} \label{eq:source_lce} L_{\mbox{ce}}^{S} = -\frac{1}{\mathrm{D}^{S}} \sum_{i=1}^{\mathrm{D}^{S}} 
[(C_S(F_\theta(X^{S}_{i})), Y^{S}_{i})]
\end{equation} 
\noindent where $F_{\theta}$ is the feature extractor, $C_S(.)$ is the source classifier, $D^S$ is the total number of source domains, $X^{S}_i$ and ${Y^{S}_i}$ are the image-label pair for samples in the $i^{th}$ source subject domain. The overall source subject loss is expressed as: 
% $\mathcal{L}_{combine}^S = \mathrm{\mathcal{L}}_{ce}^{S} + \mathcal{L}_{mmd} [{MMD}(N_{}^{S}, N^{\widetilde{S}})]$
\begin{equation}
    L^{S} = L_{\mbox{ce}}^{S} + \lambda L_{\mbox{mmd}}^{S}
\end{equation}
\noindent where $L_{\mbox{ce}}^{S}$ is the supervised loss, $L_{\mbox{mmd}}^{S}$ is the domain discrepancy loss to align source subjects in a common feature space, and $\lambda$ weights their contributions.

%%%%%%%%%%%%%
\subsection{Target Adaptation with Augmented Confident Target Pseudo-Labels}  

After training the source classifier and aligning the subject domains, the next step is generating PLs for the target domain. However, generating target labels using the trained source classifier ($C_S$) produced noisy PLs \cite{deng2022robust}. Thus, we proposed an ACPL strategy to produce a reliable PL using augmentation thresholding before the adaptation. For each target sample $x^{T}_j$, we apply a single augmentation to produce an augmented version of the sample $\widehat{x}^{T}_j$. Using the trained source subject classifier $C_S$ we make predictions by producing softmax probability ${p_j}=\mbox{softmax}[C_S(x^T_j)]$ and $\widehat{p_j}=\mbox{softmax}[C_S(\widehat{x}^T_j)]$. The probability dictionary for every example and its augmented version is defined as:

% We introduce the target subjects $X^T_n$, apply a augmentation to produce $\widehat{X}^{T}_n$ an augmented version of each data sample. 
% % Thus we denote $X^{T}$ as $X^{T}_{org}$. 
% Using the trained source subject classifier $C_S$ we make predictions by producing softmax probability $\textbf{P}^T_n=\mbox{softmax}[C_S(X^T_n)]$ and $\widehat{\textbf{P}}^T_n=\mbox{softmax}[C_S(\widehat{X}^T_n)]$ of every example and its augmented version to define the probability dictionaries:
% and $\widehat{P}^T=\sigma[F_\theta(\widehat{X}^T)]$ 
% The source-trained classifier $C_S$ is used to produce the logits that were then passed to softmax to produce probabilities of the $X^{T}_{org}$ and $X^{T}_{aug}$ samples:
% \begin{equation}
% \left\{\begin{matrix}
% P^T_{org}= \sigma[C_S(X^T_{org})]\\ \\
% P^T_{aug}= \sigma[C_S(X^T_{aug})]
% \end{matrix}\right.
% \end{equation}
{\small
\begin{equation}
\textbf{P}^T_n = \left\{p_1, ..., p_n\right\}, \quad  
\widehat{\textbf{P}}^T_n= \left\{\widehat{p}_1, ..., \widehat{p}_n\right\}
\end{equation}}
\noindent where $\textbf{P}^T_n$ and $\widehat{\textbf{P}}^T_n$ are the probabilities of the original and augmented version of every target sample. We calculate the average of these probabilities; $a_n = \{(p_1+ \hat{p}_1) / 2, ...,  (p_n+ \hat{p}_n) / 2\}$, where $a_n$
% $\textbf{A}^T_n = (\textbf{P}^T_n+ \widehat{\textbf{P}}^T_n) / 2$
%and define $\textbf{A}^T_n = \left\{ a_1, ..., a_n \right\}$. 
% The $\sigma$ is the softmax that converts the logits into $P^T_{org}$ and $P^T_{aug}$ and take the average of these probabilities:
% \begin{equation}
% \textbf{A}^T_n = \left\{ avg(p_1, \widehat{p}_1), ..., avg(p_n, \widehat{p}_n) \right\}
% \end{equation}
% These probabilities of each version of the sample are then used to take the average of the probabilities and if the value is greater than the threshold $\alpha$ we consider that a reliable sample. Now we have two labels for each sample, In an ideal case the labels have to be similar but because of the subtle nature of the expression in pain estimation, the labels tend to differ from each other. 
%It represents a dictionary containing the average probability of every sample in the target domain. 
undergo thresholding to consider samples with a value greater than $\tau$. To assign a PLs ${\widehat{Y}_{n}^{T} = \{\hat{y}_1, ..., \hat{y}_n\}}$ to the samples, we choose the label of the sample version $\textbf{P}^T$ or $\widehat{\textbf{P}}^T$ which has the highest probability $p^*_j = max(p_j, \hat{p}_j)$, where max is an element-wise operator and consider that the label for the target sample defined as: 
% \begin{equation}
%     \widehat{\textbf{Y}}^T_n = \arg \max_{i\in n} \left\{ (p_i, \widehat{p}_i) \cdot \mathbb{I} {(a_i>\tau)}  \right\}
% \end{equation} 
\begin{equation}
    \widehat{\textbf{Y}}^T_n = \arg \max_{j\in n} \left\{ (p^*_j) \cdot \mathbb{I} {(a_j>\tau)}  \right\}
\end{equation} 
%\noindent where $H^T$ contains the probability of the class that we assign to the target samples.  
% \noindent where $p^*_i$ and $\widehat{p}_i$ are the two separate probabilities, 
\noindent where $p^*_j$ is the max probability and $a_j$ is the average probability of $p_j$ and $\widehat{p}_j$, which is the $j^{th}$ sample in the target subject. We take the $\arg\max$ for the samples whose average probability $a_j$ is greater than $\tau$ using the $\mathbb{I}$ function, assuring that only the confident samples are chosen for the target domain.
%Then, they are converted into  hard labels $\widehat{\textbf{Y}}^T$: 
% \begin{equation}
%     \widehat{\textbf{Y}}^T = \mbox{argmax}(H^T)
% \end{equation}
% \begin{equation}
% \left\{\begin{matrix}
% (x_{i}^{t},\hat{y}_{i}^{t}) \epsilon \mathit{N^T_{cl}}, &if \max_{t, t'} avg (p_{i}(x_{i}^{t}), p_{i}(x_{i}^{t'}) ) > \alpha \\ &otherwise
% \end{matrix}\right.
% \end{equation}
%\noindent and converting them into hard labels $\widehat{\textbf{Y}}^T$. 
%\noindent $\mathbb{I}$ defines the indicator function, assuring that only the confident samples are chosen for the target domain. 
${\widehat{Y}_{n}^{T}}$ represents the target PLs, defining the confident target subject domain as:
\begin{equation}
    \mathit{D^T_{cl}}= (X_{}^{T},\widehat{Y}_{}^{T}) 
\end{equation}
% which has the  $x_i^t$ and $x_i^{t'}$ is the original and augmented version of the sample. $max_{t,t'}$ will consider the label which has the highest probability.

\noindent The same source feature extractor $F_\theta$ is now used for the target subject adaptation. Both source $D^S$ and target subjects $D_{cl}^{T}$ are introduced, and for the target, we only include the confident samples. To align the source subjects with the target, the MMD loss is calculated using Eq.~\ref{eq:main_mmd} between all the aligned subject domains ($D^S$) and the target ($D_{cl}^{T}$), where $F^S = F_{\theta}(X^{S}_i)$ and $F^T= F_{\theta}(X^T_n)$ respectively:
% {\small
% \begin{equation} \label{eq:train_mmd} 
% {MMD}(N_{}^{S}, N_{cl}^{T})=
% \\ \mathrm{\left\| \mathbb{{E}}_{X_{}^{S}} [F_\theta(\mathrm{X}^{S})]-\mathbb{{E}}_{X_{}^{T}} [F_\theta(\mathrm{X}^{T})] \right\|}_{\mathcal{H}}^{2} 
% \end{equation}}
\begin{equation} 
     L_{\mbox{mmd}}^{S,T} = {MMD}(F^{S}, F^T)
\end{equation}
% \begin{equation}
% {{Y_{n}^{T}}}=\left \{\hat{y_{1}}, \hat{y_{2}}, ..., \hat{y_{n}} \right \}
% \end{equation}
\noindent where $F^S$ is the combination of all the source subjects we already aligned using Eq.~\ref{eq:src_mmd}, now reducing the discrepancy with $F^T$ that consists of confident target distributions. The supervised loss is also calculated for the target subject with confident PLs:

\begin{equation} \label{eq:tar_lce}  L_{\mbox{ce}}^{T}  = -\frac{1}{\mathrm{D}^{T}_{cl}} \sum_{j=1}^{\mathrm{D}^{T}_{cl}}
C_T(F_\theta(x^T_{j}), \widehat{y}_j)
\end{equation}
\noindent where the target classifier $C_T$ is trained on target confident samples $x^T_j$ with PLs $\widehat{y}_j$. The PL is updated after every $M$ epoch. Finally, we define the overall objective function for the adaptation to the target subject as follows:
\begin{equation} \label{eq:target_loss} 
L_{\mbox{adapt}}^{T} = L_{\mbox{ce}}^{S} + L_{\mbox{ce}}^{T} + L_{\mbox{mmd}}^{S,T}
\end{equation}
Here, we minimize the target adaptation loss $L_{\mbox{adapt}}^{T}$, containing discrepancy loss between source and target subjects $L_{\mbox{mmd}}^{S,T}$ and their supervision losses $L_{\mbox{ce}}^{S}$ (defined in Eq. \ref{eq:source_lce}) and $L_{\mbox{ce}}^{T}$ respectively. $L_{\mbox{adapt}}^{T}$ was minimized in a mini-batch manner until the model convergences.

\begin{table*}
%\scriptsize
\renewcommand{\arraystretch}{1.4}
    \centering
    %\caption{ \textbf{Comparison of our \emph{Subject-Based Domain Adaptation} method with source-only and STA MSDA methods on 77 labeled sources adapted to 10 unlabeled target subjects.}}
    \caption{Accuracy of our Subject-based MSDA and state-of-the-art methods on BioVid for 10 target subjects with all 77 sources. \textbf{Bold} text shows the highest and \textit{Italic} shows the second best accuracy.}
    \label{tab:src77_comparison_srccombined_stamsda}
   \begin{tabular}{c|c||c|c|c|c|c|c|c|c|c|c|c}
\thickhline
\textbf{Setting}       & \textbf{Methods}                                                                     & \textbf{Sub-1}                                               & \textbf{Sub-2}                                               & \textbf{Sub-3}                                               & \textbf{Sub-4}                                               & \textbf{Sub-5}                                               & \textbf{Sub-6}                                               & \textbf{Sub-7}                                               & \textbf{Sub-8}                                               & \textbf{Sub-9}                                               & \textbf{Sub-10}                                              & \textbf{Avg}                       \\ \hline \thickhline
\textbf{Source combined} & \begin{tabular}[c]{@{}c@{}}Source-only\\ UDA (Ours)\end{tabular}       & \begin{tabular}[c]{@{}c@{}}0.62\\ 0.73\end{tabular} & \begin{tabular}[c]{@{}c@{}}0.61\\ 0.64\end{tabular} & \begin{tabular}[c]{@{}c@{}}0.65\\ 0.73\end{tabular} & \begin{tabular}[c]{@{}c@{}}0.55\\ 0.59\end{tabular} & \begin{tabular}[c]{@{}c@{}}0.51\\ 0.54\end{tabular} & \begin{tabular}[c]{@{}c@{}}0.71\\ 0.75\end{tabular} & \begin{tabular}[c]{@{}c@{}}0.70\\ 0.76\end{tabular} & \begin{tabular}[c]{@{}c@{}}0.52\\ 0.53\end{tabular} & \begin{tabular}[c]{@{}c@{}}0.54\\ 0.51\end{tabular} & \begin{tabular}[c]{@{}c@{}}0.55\\ 0.58\end{tabular} & \begin{tabular}[c]{@{}c@{}}0.59\\ 0.63\end{tabular} \\ \hline \hline
\textbf{Multi-source} & \begin{tabular}[c]{@{}c@{}}M\textsuperscript{3}SDA \cite{peng2019moment}\\ CMSDA \cite{scalbert2021multi}\\ SImpAI \cite{venkat2020your}\\ MSDA (Ours)\\ MSDA (Ours)\textsubscript{top-k} \end{tabular} & \begin{tabular}[c]{@{}c@{}}0.67\\ \textbf{0.93} \\ \emph{0.80}\\ \textbf{0.93}\\\textbf{0.93}\end{tabular} & \begin{tabular}[c]{@{}c@{}}0.66\\0.47\\ \emph{0.69}\\ \emph{0.69} \\ \textbf{0.71}\end{tabular} & \begin{tabular}[c]{@{}c@{}}0.61\\0.81\\0.55\\ \emph{0.84}\\ \textbf{0.86} \end{tabular} & \begin{tabular}[c]{@{}c@{}}0.58\\ \textbf{0.87} \\ \emph{0.75}\\ 0.66\\ \textbf{0.87}\end{tabular} & \begin{tabular}[c]{@{}c@{}}0.55\\0.53\\0.52\\ \emph{0.60}\\ \textbf{0.88}\end{tabular} & \begin{tabular}[c]{@{}c@{}}0.50\\ \emph{0.84} \\0.81\\ 0.76\\ \textbf{0.92}\end{tabular}  & \begin{tabular}[c]{@{}c@{}}0.67\\0.57\\0.71\\ \emph{0.84}\\\textbf{0.86}\end{tabular} & \begin{tabular}[c]{@{}c@{}}0.56\\0.54\\ \emph{0.61}\\ 0.55\\\textbf{0.77}\end{tabular} & \begin{tabular}[c]{@{}c@{}}0.54\\ \emph{0.74} \\ 0.59\\ 0.62\\\textbf{0.84}\end{tabular} & \begin{tabular}[c]{@{}c@{}}0.67\\ \textbf{0.70} \\0.56\\ 0.66\\ \emph{0.68}\end{tabular} & \begin{tabular}[c]{@{}c@{}}0.60\\0.70\\0.65\\ \emph{0.71}\\ \textbf{0.83}\end{tabular} \\ \hline \hline
\textbf{Oracle}          & Fully-Supervised                                                                  & 0.99                                                & 0.91                                                & 0.98                                                & 0.97                                                & 0.98                                                & 0.97                                                & 0.96                                                & 0.95                                                & 0.99                                                & 0.98                                                & 0.96                                               \\ \thickhline
\end{tabular}
     %\vspace{-8mm}
% ALEKOE
\vspace{-10pt}
\end{table*}

%%%%%%%%%%%%
\section{Results and Discussion}

%%%%%%
\subsection{Experimental Methodology}

\noindent \textbf{Datasets} 
% We follow the standard protocol for measuring the performance of our multi-subjects domain adaptation for pain estimation. We use the BioVid heat and pain dataset [] to determine the efficacy of our model, 
In our proposed \emph{subject-based domain adaptation} for FER, as we consider each subject as a domain, it is essential to use a dataset that includes comprehensive subject-related information and encompasses a diverse array of individuals. To determine the efficacy of our model, we use the BioVid heat and pain(PartA) \cite{walter2013biovid} and UNBC-McMaster shoulder pain datasets \cite{lucey2011painful}. The BioVid (PartA) consists of 87 subjects captured in a controlled environment, where each subject falls into one of five categories: "no pain" and four distinct pain levels denoted as PA1, PA2, PA3, and PA4, from lower to higher.
Previous research \cite{werner2017analysis} has shown that the initial pain intensities failed to show any facial activities and suggested focusing only on no pain and the highest pain intensities. Our experiments consider two classes: no pain and pain level 4 (PA4). Each subject consists of 20 videos per class, and each video lasts 5.5s. Based on \cite{werner2017analysis}, PA4 does not show any facial activity in the first 2s of the video. Therefore, we only consider frames after 2s to eliminate the initial part of the sequence that did not indicate any response to the pain. The UNBC-McMaster dataset consists of 25 subjects with 200 video sequences. The pain intensity for each frame in the video is scored using the PSPI scale \cite{prkachin1992consistency}, which ranges from 0 to 15. Due to the significant imbalance between pain intensities, we are following the same quantization strategy as \cite{rajasekhar2021deep, ruiz2018multi} that divides the pain intensities into five discrete levels: 0 (no pain), 1(1), 2(2), 3(3), 4(4-5), 5(6-15).

\noindent \textbf{Experimental Protocol.}
To evaluate the performance of our method, we define subjects as source and target domains. The first experiment was conducted on the BioVid dataset, where 77 subjects were treated as sources and adapted to the remaining ten target subjects. The following experiment is on the UNBC-McMaster dataset, which includes 20 subjects in the source domain adapted to the remaining five subjects in the target domain. To define a baseline for \emph{subject-based domain adaptation} for the recognition of pain, we follow the standard protocol defined for MSDA methods \cite{xu2018deep, peng2019moment}.
\textbf{Source-combined:} We first define the lower-bound, which is the traditional approach of training a model by using all the sources and testing the target subject. This approach is also known as source-only as it does not adapt to the target data. The second experiment combines all subjects as before and then adapts to a target subject as in standard UDA. %We perform UDA in our subject-based domain adaptation method by combining subject IDs in a single source and adapting them to the target subject. 
\textbf{Multi-source DA:} We treat each subject as a separate domain and reduce the domain shift among the different sources before adapting to the target.
To the best of our knowledge, this is the first work introducing MSDA to FER. No method is directly comparable. Hence, we evaluate our method with three commonly used MSDA approaches: M\textsuperscript{3}SDA \cite{peng2019moment}, contrastive multi-source domain adaptation (CMSDA) \cite{scalbert2021multi} and self-supervised implicit alignment (SImpAl) \cite{venkat2020your}. The M\textsuperscript{3}SDA technique was the baseline method for MSDA classification tasks that reduced the discrepancy based on the moment-matching approach between domains. CMSDA and SImpAI methods are based on generating the target PLs. These models used a single feature extractor for the source and target adaptation, which reduces the cost of training multiple networks, avoiding the computational complexity when dealing with many domains. \textbf{Oracle:} It is the upper bound where we fine-tune the source model by leveraging labels of every target image in a fully supervised manner.

% \begin{table*}[]
% \renewcommand{\arraystretch}{1.4}
%     \centering
%     \caption{ \textbf{Performance of $MS^bDA$ when the number of sources is 30, 60, and 77 on 10 unlabeled target subjects.}}
%     \label{tab:src30_60_77_comparison}
% \begin{tabular}{|c|c|c|c|c|c|c|c|c|c|c|c|c|}
% \hline
% \multirow{4}{*}{\begin{tabular}[c]{@{}c@{}}\textbf{Multi-source}\\ \textbf{DA}\end{tabular}} & \textbf{Sources (sub)} & \textbf{Sub-1} & \textbf{Sub-2} & \textbf{Sub-3} & \textbf{Sub-4} & \textbf{Sub-5} & \textbf{Sub-6} & \textbf{Sub-7} & \textbf{Sub-8} & \textbf{Sub-9} & \textbf{Sub-10} & \textbf{Avg}  \\ \cline{2-13} 
%                                                                            & \textbf{30}      & 0.71  & 0.63  & 0.50  & 0.56  & 0.70  & 0.56  & \textbf{0.82}  & 0.52  & \textbf{0.84}  & 0.54   & 0.63 \\ \cline{2-13} 
%                                                                            & \textbf{60}      & \textbf{0.94}  & 0.61  & 0.65  & \textbf{0.67}  & \textbf{0.89}  & \textbf{0.91}  & \textbf{0.82}  & \textbf{0.58}  & 0.54  & \textbf{0.62}   & \textbf{0.72} \\ \cline{2-13} 
%                                                                            & \textbf{77}      & 0.93  & \textbf{0.69}  & \textbf{0.84}  & 0.64  & 0.57  & 0.85  & 0.81  & \textbf{0.58}  & 0.60  & 0.60   & 0.71 \\ \hline
% \end{tabular}
% \end{table*}

\noindent \textbf{Implementation Details.}
In all experiments, we use the ResNet18 \cite{he2016deep} backbone pre-trained on ImageNet by removing the first ReLU and MaxPool layers and the last 2D adaptive average pooling layer. In our experiments, the backbone is shared across all the source domains, followed by the shared classifier. The images are resized to 100$\times$100 resolution, and the model is trained with stochastic gradient descent (SGD) with a batch size of 16 and a learning rate of $10^{-4}$. We initially set a high threshold value of $\tau=0.90$ for the PL selection because, at the start of the training, the samples tended to be more noisy, but as training progressed, we decreased the value by 0.02 after every 20 epochs. For our ACPL technique to generate reliable PLs for the target domain, we experimented with different augmentation techniques such as horizontal flip, rotation \ang{90}, sharpness, and vertical flip. Among them, horizontal flip outperforms other methods and produces the most confident PL. Therefore, we use the horizontal flip as an augmentation technique in all our experiments.
\subsection{Comparison with the State-of-the-Art}
The experiments on the BioVid dataset are reported in Table \ref{tab:src77_comparison_srccombined_stamsda}. In the source-combine source-only method, we follow the lower-bound approach without any form of domain adaptation, where a model is trained on training subjects (sources) and evaluated on distinct target subjects. As expected, all the other techniques outperform the source-only approach and verify our hypothesis that there can be an important domain gap between different identities in the same dataset. We then perform UDA on our subject-based approach, where all the subjects are combined, treated as a single source, and then adapted to the target subject. This simple adaptation already improves the method's accuracy by 0.4. However, this is not considered if there is a domain shift between source subjects. For example, if some subjects are similar and others have some dissimilarities based on their cultural or ethnic difference, simply combining them is not optimal. Therefore, MSDA approaches are introduced to deal with the discrepancy between the subjects. 

We compare our subject-based MSDA to state-of-the-art MSDA methods. The M\textsuperscript{3}SDA methods have a slight performance gain over source-only 0.1. The SImpAI and CMSDA have significantly improved over source-only with 0.6 and 0.11, respectively, whereas our approach outperforms all methods with an overall accuracy of 0.71.
Nonetheless, it can be observed that there is variability across different target subjects' accuracy in the subject-based MSDA, e.g., \emph{Sub-1} has an accuracy of {0.93}, and \emph{Sub-8} has an accuracy of {0.55}. These differences highlight the significance of selecting relevant source subjects for every target domain to achieve higher adaptability. Therefore, the top 30 closest sources were selected for every target domain. Specifically, we calculate MMD distances between each target and every source subject and pick the top 30 closest sources for the respective target subject to utilize in the adaptation process. It improves the results of all target subjects and reaches an average accuracy of 0.83, surpassing all the methods. In a fully-supervised (Oracle) approach, we use ResNet18 as a backbone, performing fine-tuning using labeled target images, achieving an average accuracy of 0.96.
% Additional experimental details for top k and experiments with different numbers of source subjects are presented in Supplementary material\footnote{\textcolor{red}{\textbf{Supplementary material} contains our experiments with different numbers of source subjects}}. 

Table \ref{tab:src77_comparison_srccombined_unbc_results} shows the performance of our proposed method on the UNBC-McMaster should pain dataset. Our approach outperforms all methods on each target domain, including source-only and state-of-the-art MSDA methods, achieving an average accuracy of 0.86. More specifically, subject-based MSDA improves the performance by 0.6 compared to M\textsuperscript{3}SDA, and by 0.5 over CMSDA and SImpAI approaches. We did not perform a top-k experiment for this dataset due to a limited number of source subjects, i.e., 20.

\begin{table}
\renewcommand{\arraystretch}{1.4}
\centering
\caption{Accuracy of Subject-based MSDA with baseline methods on \textbf{UNBC-McMaster shoulder pain} by selecting five subjects as targets and the remaining 20 subjects as source domains.}
\begin{tabular}{c||c|c|c|c|c|c}
\thickhline
 \textbf{Methods}                                                                         & \textbf{Sub-1}                                                          & \textbf{Sub-2}                                                          & \textbf{Sub-3}                                                          & \textbf{Sub-4}                                                          & \textbf{Sub-5}                                                          & \textbf{Avg}                                                            \\ \thickhline
 \begin{tabular}[c]{@{}c@{}}Source-only\\ UDA (Ours)\end{tabular}             & \begin{tabular}[c]{@{}c@{}}0.74\\ 0.76\end{tabular}            & \begin{tabular}[c]{@{}c@{}}0.84\\ 0.87\end{tabular}            & \begin{tabular}[c]{@{}c@{}}0.81\\ 0.84\end{tabular}            & \begin{tabular}[c]{@{}c@{}}0.68\\ 0.70\end{tabular}            & \begin{tabular}[c]{@{}c@{}}0.83\\ 0.85\end{tabular}            & \begin{tabular}[c]{@{}c@{}}0.78\\ 0.80\end{tabular}            \\ \hline \hline 
 \begin{tabular}[c]{@{}c@{}}M\textsuperscript{3}SDA \cite{peng2019moment}\\ CMSDA \cite{scalbert2021multi}\\ SImpAI \cite{venkat2020your}\\ MSDA (Ours)\end{tabular} & \begin{tabular}[c]{@{}c@{}}0.78\\ 0.80\\ 0.80\\ \textbf{0.81}\end{tabular} & \begin{tabular}[c]{@{}c@{}}0.87\\ 0.86\\ 0.88\\ \textbf{0.91}\end{tabular} & \begin{tabular}[c]{@{}c@{}}0.92\\ 0.83\\ 0.81\\ \textbf{0.94}\end{tabular} & \begin{tabular}[c]{@{}c@{}}0.66\\ 0.71\\ 0.70\\ \textbf{0.72}\end{tabular} & \begin{tabular}[c]{@{}c@{}}0.81\\ 0.85\\ 0.87\\ \textbf{0.92}\end{tabular} & \begin{tabular}[c]{@{}c@{}}0.80\\ 0.81\\ 0.81\\ \textbf{0.86}\end{tabular} \\ \hline \hline
 \begin{tabular}[c]{@{}c@{}}Oracle\end{tabular}                                                               & 0.96                                                           & 0.98                                                           & 0.97                                                           & 0.94                                                           & 0.97                                                           & 0.96                                                           \\ \thickhline
\end{tabular}
\label{tab:src77_comparison_srccombined_unbc_results}
\end{table}

\begin{figure}
\hspace{0mm}
\centering
\includegraphics[width=0.47\textwidth]{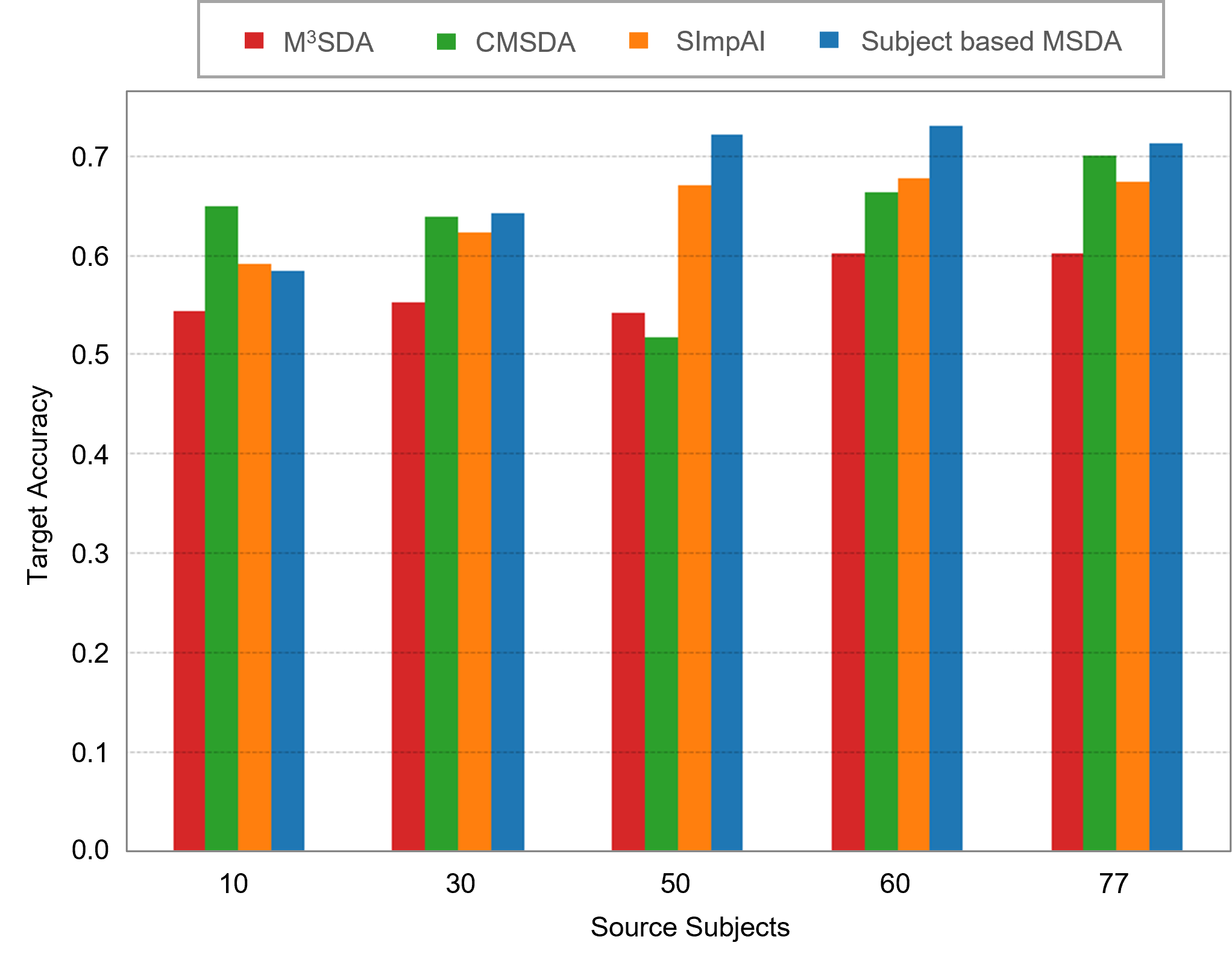}
\caption{Performance comparison of \emph{Subject based MSDA, SImpAI, CMSDA}, and \emph{M\textsuperscript{3}SDA} when the number of source subjects is 10, 30, 50, 60, and 77 adapted to 10 unlabeled target subjects.}
\label{fig:incre_num_subs}
% ALEKOE
\vspace{-10pt}
\end{figure}

\begin{figure*}
\hspace{0mm}
\centering
\includegraphics[width=0.95\textwidth]{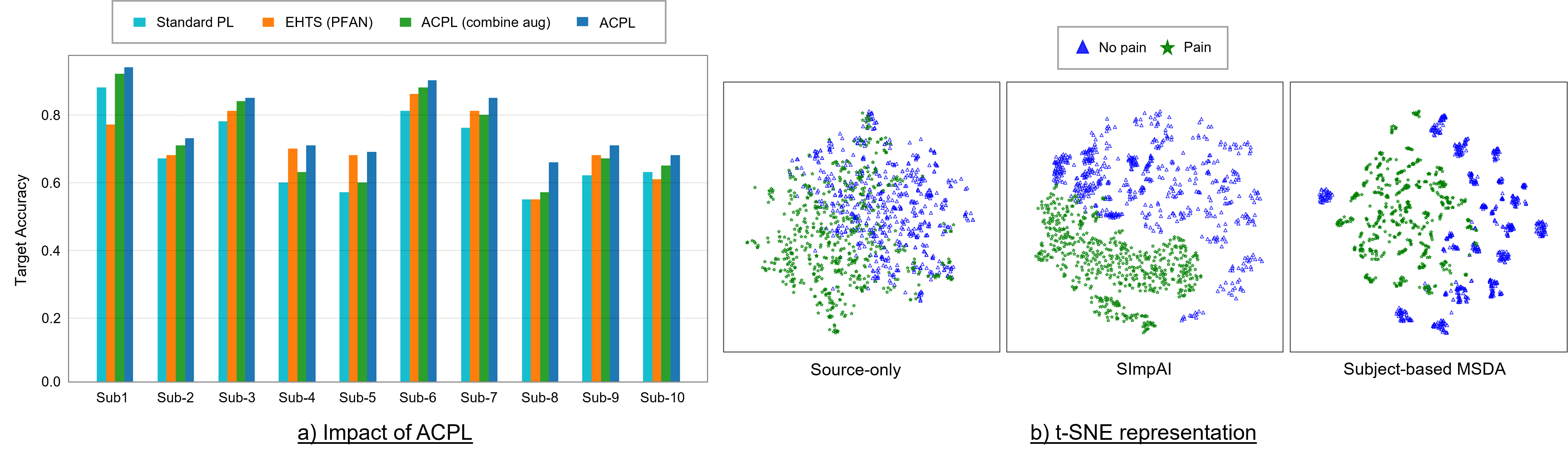}
\caption{\textbf{(a)} Comparison between techniques for generating target PLs. The \emph{cyan bar} indicates a standard way of generating PL, while the \emph{orange bar} indicates the EHTS approach. Our ACPL strategy by combining different augmentation, i.e., horizontal-vertical-flip, increase sharpness, and rotation \ang{90}, is shown in \emph{green bar}. Finally, the ACPL technique with only horizontal-flip augmentation is indicated by \emph{the blue line}. \textbf{(b)} A t-SNE projection of embeddings from source to target subjects \textbf{SUB-1}. a) represents the source-only (baseline) setting without adaptation. b) shows the representation generated from the SImpAI method. c) represents our approach feature embeddings for the target subject. Color and shape represent different classes, Blue for \textit{No-Pain} and green for \textit{Pain}. (Best viewed in color.)}.
\label{fig:pl_tsne}
\end{figure*}

%%%%%
\subsection{Ablation Studies}

\noindent \textbf{Impact of the number of source subjects.}
% In Table \ref{tab:src60_comparison_srccombined_stamsda}, we have shown the increase in target subject adaptability rate when subjects increased to 60 by 0.8\%. 
This section analyzes the impact on model adaptation when the number of subjects grows in the Biovid dataset. The experiments were performed with 10, 30, 50, 60, and 77 source subjects individually adapted to 10 unlabeled target subjects on \emph{Subject-based MSDA}, SImpAI, CMSDA, and M\textsuperscript{3}SDA methods. Fig. \ref{fig:incre_num_subs} illustrates the average performance of all target models. 
We see the trend of increase in target adaptability when more source subjects were added except for CMSDA, whose performance dropped on \emph{Sub-30} and \emph{Sub-50}. Our method outperforms with an increasing number of subjects, which shows that our model is more adaptable with more source domains.
% All the methods have the highest accuracy gain when there are 60 source subjects except for CMSDA, which achieves its highest peak with 77 source subjects. 
By increasing the number of sources, the model tends to learn different variations by adding more diverse subjects that include a variety of individuals with diverse ways of expressing pain. When sources were increased to 77, the model did not improve but had a slight decrease in accuracy \emph{Subject-based MSDA} and relatively similar performance in SImpAI and {M\textsuperscript{3}SDA} as compared to the model train with 60 subjects. This shows that the model is saturated and does not learn any new subjective variability when more subjects are added beyond 60. 
Furthermore, we have shown in Section IV (B) that selecting the sources \textit{(top-k)} that are identical for the given target subject has a much more significant impact on the model convergence. This highlights the need to increase the number of related source subjects.

\noindent \textbf{Impact of augmented confident PL.}
We compare the effectiveness of our ACPL generation technique with three other methods to generate target PLs on the Biovid dataset. In this experiment, we have to use the target labeled information to estimate the reliability of the generated PLs by comparing them with the ground truth. For this diagnosis, we are considering a model trained on 77 sources and then introducing ten fixed, unlabeled target subjects illustrated in Fig.~\ref{fig:pl_tsne}a. First, the standard way to produce the target PL is directly from the source classifier without a confidence threshold. Next, the "Easy-to-Hard Transfer Strategy" \cite{chen2019progressive} applies a PL to the target instance based on the nearest source class cluster, which includes easy examples first and gradually adds more images to the adaptation process as the training progresses.
Furthermore, we also evaluated ACPL by combining different augmentations, including horizontal-vertical-flip, increased sharpness, and rotation \ang{90} for generating PLs. Finally, the ACPL is applied to each target subject, where we use two versions of each sample, an original and a horizontal flip, as an augmented image. It can be noted in Fig.~\ref{fig:pl_tsne}a that ACPL improves the target PL in all subjects with an average accuracy of 0.77, indicating the reliability of target samples before the target adaptation.

\subsection{Visualization of Feature Distributions}
To have a better insight into our proposed method, the target feature embeddings of the Biovid dataset are visualized using a t-SNE \cite{van2008visualizing} graph. Our method is compared with Baseline and SImpAI \cite{venkat2020your} method, i.e., source-only and STA MSDA method illustrated in Fig.~\ref{fig:pl_tsne}b. It can be seen that the feature representation in source-only for pain and no-pain classes is cluttered, not aligned, and category-unaware. Compared to SImpAI, the categories are much more separable from each other, but still, they were not aligned. However, our \emph{Subject-based MSDA} method takes advantage of discrepancy loss to align the target features of the same video much closer to each other, which shows the benefit of reducing the disparity in the target feature alignment process.

\section{Conclusions}

In this paper, we proposed the \emph{Subject-based MSDA} technique to incorporate subjects as multiple domains in FER. As subjects have differences in skin color, facial cues, and facial expressions, it makes sense to consider each subject as a different domain and adapt to that in a multi-domain setting. 
%We introduced a new approach of including subject IDs as domains contrary to each dataset as a domain in a standard domain adaptation setting. 
In addition, we proposed a technique to generate reliable target PLs using ACPL before the target adaptation. Extensive experiments have shown that our approach for dealing with many source subjects in target subject adaptation performs better than source-only (no adaptation), as in regular FER training and single UDA approaches. Furthermore, we also compared our technique with the current MSDA methods and achieved higher performance. Finally, we showed that increasing the number of source subjects improves the target performance. However, selecting the relevant source subjects has a much higher impact on the adaptation process (Biovid). In the future, we aim to include more individuals from different datasets to increase the robustness of the target model and perform cross-corpus evaluation.

\noindent \textbf{Acknowledgments:} This work was supported in part by the Fonds de recherche du Québec – Santé (FRQS), the Natural Sciences and Engineering Research Council of Canada (NSERC), Canada Foundation for Innovation (CFI), and the Digital Research Alliance of Canada.

\balance
{\small
\bibliographystyle{ieee}
\bibliography{egbib}

\begin{thebibliography}{10}\itemsep=-1pt

\bibitem{barrett2007experience}
L.~F. Barrett, B.~Mesquita, K.~N. Ochsner, and J.~J. Gross.
\newblock The experience of emotion.
\newblock {\em Annu. Rev. Psychol.}, 58:373--403, 2007.

\bibitem{ben2010theory}
S.~Ben-David, J.~Blitzer, K.~Crammer, A.~Kulesza, F.~Pereira, and J.~W. Vaughan.
\newblock A theory of learning from different domains.
\newblock {\em Machine learning}, 79:151--175, 2010.

\bibitem{bozorgtabar2020exprada}
B.~Bozorgtabar, D.~Mahapatra, and J.-P. Thiran.
\newblock Exprada: Adversarial domain adaptation for facial expression analysis.
\newblock {\em Pattern Recognition}, 100:107111, 2020.

\bibitem{chen2019progressive}
C.~Chen, W.~Xie, W.~Huang, Y.~Rong, X.~Ding, Y.~Huang, T.~Xu, and J.~Huang.
\newblock Progressive feature alignment for unsupervised domain adaptation.
\newblock In {\em Proceedings of the IEEE/CVF conference on computer vision and pattern recognition}, pages 627--636, 2019.

\bibitem{chen2021cross}
T.~Chen, T.~Pu, H.~Wu, Y.~Xie, L.~Liu, and L.~Lin.
\newblock Cross-domain facial expression recognition: A unified evaluation benchmark and adversarial graph learning.
\newblock {\em IEEE transactions on pattern analysis and machine intelligence}, 2021.

\bibitem{deng2022robust}
Z.~Deng, D.~Li, Y.-Z. Song, and T.~Xiang.
\newblock Robust target training for multi-source domain adaptation.
\newblock {\em arXiv preprint arXiv:2210.01676}, 2022.

\bibitem{ghifary2016deep}
M.~Ghifary, W.~B. Kleijn, M.~Zhang, D.~Balduzzi, and W.~Li.
\newblock Deep reconstruction-classification networks for unsupervised domain adaptation.
\newblock In {\em European conference on computer vision}, pages 597--613. Springer, 2016.

\bibitem{goodfellow2013challenges}
I.~J. Goodfellow, D.~Erhan, P.~L. Carrier, A.~Courville, M.~Mirza, B.~Hamner, W.~Cukierski, Y.~Tang, D.~Thaler, D.-H. Lee, et~al.
\newblock Challenges in representation learning: A report on three machine learning contests.
\newblock In {\em International conference on neural information processing}, pages 117--124. Springer, 2013.

\bibitem{gretton2012kernel}
A.~Gretton, K.~M. Borgwardt, M.~J. Rasch, B.~Sch{\"o}lkopf, and A.~Smola.
\newblock A kernel two-sample test.
\newblock {\em The Journal of Machine Learning Research}, 13(1):723--773, 2012.

\bibitem{han2020personalized}
J.~Han, L.~Xie, J.~Liu, and X.~Li.
\newblock Personalized broad learning system for facial expression.
\newblock {\em Multimedia Tools and Applications}, 79(23):16627--16644, 2020.

\bibitem{he2016deep}
K.~He, X.~Zhang, S.~Ren, and J.~Sun.
\newblock Deep residual learning for image recognition.
\newblock In {\em Proceedings of the IEEE conference on computer vision and pattern recognition}, pages 770--778, 2016.

\bibitem{hu2018squeeze}
J.~Hu, L.~Shen, and G.~Sun.
\newblock Squeeze-and-excitation networks.
\newblock In {\em Proceedings of the IEEE conference on computer vision and pattern recognition}, pages 7132--7141, 2018.

\bibitem{kang2020contrastive}
G.~Kang, L.~Jiang, Y.~Wei, Y.~Yang, and A.~Hauptmann.
\newblock Contrastive adaptation network for single-and multi-source domain adaptation.
\newblock {\em IEEE transactions on pattern analysis and machine intelligence}, 44(4):1793--1804, 2020.

\bibitem{kitayama2006cultural}
S.~Kitayama, B.~Mesquita, and M.~Karasawa.
\newblock Cultural affordances and emotional experience: socially engaging and disengaging emotions in japan and the united states.
\newblock {\em Journal of personality and social psychology}, 91(5):890, 2006.

\bibitem{kollias2019deep}
D.~Kollias, P.~Tzirakis, M.~A. Nicolaou, A.~Papaioannou, G.~Zhao, B.~Schuller, I.~Kotsia, and S.~Zafeiriou.
\newblock Deep affect prediction in-the-wild: Aff-wild database and challenge, deep architectures, and beyond.
\newblock {\em International Journal of Computer Vision}, 127(6):907--929, 2019.

\bibitem{kossaifi2019sewa}
J.~Kossaifi, R.~Walecki, Y.~Panagakis, J.~Shen, M.~Schmitt, F.~Ringeval, J.~Han, V.~Pandit, A.~Toisoul, B.~Schuller, et~al.
\newblock Sewa db: A rich database for audio-visual emotion and sentiment research in the wild.
\newblock {\em IEEE transactions on pattern analysis and machine intelligence}, 43(3):1022--1040, 2019.

\bibitem{li2017deeper}
D.~Li, Y.~Yang, Y.-Z. Song, and T.~M. Hospedales.
\newblock Deeper, broader and artier domain generalization.
\newblock In {\em Proceedings of the IEEE international conference on computer vision}, pages 5542--5550, 2017.

\bibitem{li2018deep}
S.~Li and W.~Deng.
\newblock Deep emotion transfer network for cross-database facial expression recognition.
\newblock In {\em 2018 24th International Conference on Pattern Recognition (ICPR)}, pages 3092--3099. IEEE, 2018.

\bibitem{li2017reliable}
S.~Li, W.~Deng, and J.~Du.
\newblock Reliable crowdsourcing and deep locality-preserving learning for expression recognition in the wild.
\newblock In {\em Proceedings of the IEEE conference on computer vision and pattern recognition}, pages 2852--2861, 2017.

\bibitem{long2015fully}
J.~Long, E.~Shelhamer, and T.~Darrell.
\newblock Fully convolutional networks for semantic segmentation.
\newblock In {\em Proceedings of the IEEE conference on computer vision and pattern recognition}, pages 3431--3440, 2015.

\bibitem{long2016unsupervised}
M.~Long, H.~Zhu, J.~Wang, and M.~I. Jordan.
\newblock Unsupervised domain adaptation with residual transfer networks.
\newblock {\em Advances in neural information processing systems}, 29, 2016.

\bibitem{lucey2011painful}
P.~Lucey, J.~F. Cohn, K.~M. Prkachin, P.~E. Solomon, and I.~Matthews.
\newblock Painful data: The unbc-mcmaster shoulder pain expression archive database.
\newblock In {\em 2011 IEEE International Conference on Automatic Face \& Gesture Recognition (FG)}, pages 57--64. IEEE, 2011.

\bibitem{mollahosseini2017affectnet}
A.~Mollahosseini, B.~Hasani, and M.~H. Mahoor.
\newblock Affectnet: A database for facial expression, valence, and arousal computing in the wild.
\newblock {\em IEEE Transactions on Affective Computing}, 10(1):18--31, 2017.

\bibitem{nguyen2021stem}
V.-A. Nguyen, T.~Nguyen, T.~Le, Q.~H. Tran, and D.~Phung.
\newblock Stem: An approach to multi-source domain adaptation with guarantees.
\newblock In {\em Proceedings of the IEEE/CVF International Conference on Computer Vision}, pages 9352--9363, 2021.

\bibitem{nisbett2003culture}
R.~E. Nisbett and T.~Masuda.
\newblock Culture and point of view.
\newblock {\em Proceedings of the National Academy of Sciences}, 100(19):11163--11170, 2003.

\bibitem{odena2017conditional}
A.~Odena, C.~Olah, and J.~Shlens.
\newblock Conditional image synthesis with auxiliary classifier gans.
\newblock In {\em International conference on machine learning}, pages 2642--2651. PMLR, 2017.

\bibitem{peng2019moment}
X.~Peng, Q.~Bai, X.~Xia, Z.~Huang, K.~Saenko, and B.~Wang.
\newblock Moment matching for multi-source domain adaptation.
\newblock In {\em Proceedings of the IEEE/CVF international conference on computer vision}, pages 1406--1415, 2019.

\bibitem{praveen2020deep}
R.~G. Praveen, E.~Granger, and P.~Cardinal.
\newblock Deep weakly supervised domain adaptation for pain localization in videos.
\newblock In {\em 2020 15th IEEE International Conference on Automatic Face and Gesture Recognition (FG 2020)}, pages 473--480. IEEE, 2020.

\bibitem{prkachin1992consistency}
K.~M. Prkachin.
\newblock The consistency of facial expressions of pain: a comparison across modalities.
\newblock {\em Pain}, 51(3):297--306, 1992.

\bibitem{rajasekhar2021deep}
G.~P. Rajasekhar, E.~Granger, and P.~Cardinal.
\newblock Deep domain adaptation with ordinal regression for pain assessment using weakly-labeled videos.
\newblock {\em Image and Vision Computing}, 110:104167, 2021.

\bibitem{ren2022multi}
C.-X. Ren, Y.-H. Liu, X.-W. Zhang, and K.-K. Huang.
\newblock Multi-source unsupervised domain adaptation via pseudo target domain.
\newblock {\em IEEE Transactions on Image Processing}, 31:2122--2135, 2022.

\bibitem{ren2015faster}
S.~Ren, K.~He, R.~Girshick, and J.~Sun.
\newblock Faster r-cnn: Towards real-time object detection with region proposal networks.
\newblock {\em Advances in neural information processing systems}, 28, 2015.

\bibitem{ruiz2018multi}
A.~Ruiz, O.~Rudovic, X.~Binefa, and M.~Pantic.
\newblock Multi-instance dynamic ordinal random fields for weakly supervised facial behavior analysis.
\newblock {\em IEEE transactions on image processing}, 27(8):3969--3982, 2018.

\bibitem{saenko2010adapting}
K.~Saenko, B.~Kulis, M.~Fritz, and T.~Darrell.
\newblock Adapting visual category models to new domains.
\newblock In {\em European conference on computer vision}, pages 213--226. Springer, 2010.

\bibitem{scalbert2021multi}
M.~Scalbert, M.~Vakalopoulou, and F.~Couzini{\'e}-Devy.
\newblock Multi-source domain adaptation via supervised contrastive learning and confident consistency regularization.
\newblock {\em arXiv preprint arXiv:2106.16093}, 2021.

\bibitem{sejdinovic2013equivalence}
D.~Sejdinovic, B.~Sriperumbudur, A.~Gretton, and K.~Fukumizu.
\newblock Equivalence of distance-based and rkhs-based statistics in hypothesis testing.
\newblock {\em The annals of statistics}, pages 2263--2291, 2013.

\bibitem{simonyan2014very}
K.~Simonyan and A.~Zisserman.
\newblock Very deep convolutional networks for large-scale image recognition.
\newblock {\em arXiv preprint arXiv:1409.1556}, 2014.

\bibitem{tzeng2014deep}
E.~Tzeng, J.~Hoffman, N.~Zhang, K.~Saenko, and T.~Darrell.
\newblock Deep domain confusion: Maximizing for domain invariance.
\newblock {\em arXiv preprint arXiv:1412.3474}, 2014.

\bibitem{van2008visualizing}
L.~Van~der Maaten and G.~Hinton.
\newblock Visualizing data using t-sne.
\newblock {\em Journal of machine learning research}, 9(11), 2008.

\bibitem{venkat2020your}
N.~Venkat, J.~N. Kundu, D.~Singh, A.~Revanur, et~al.
\newblock Your classifier can secretly suffice multi-source domain adaptation.
\newblock {\em Advances in Neural Information Processing Systems}, 33:4647--4659, 2020.

\bibitem{walter2013biovid}
S.~Walter, S.~Gruss, H.~Ehleiter, J.~Tan, H.~C. Traue, P.~Werner, A.~Al-Hamadi, S.~Crawcour, A.~O. Andrade, and G.~M. da~Silva.
\newblock The biovid heat pain database data for the advancement and systematic validation of an automated pain recognition system.
\newblock In {\em 2013 IEEE international conference on cybernetics (CYBCO)}, pages 128--131. IEEE, 2013.

\bibitem{werner2017analysis}
P.~Werner, A.~Al-Hamadi, and S.~Walter.
\newblock Analysis of facial expressiveness during experimentally induced heat pain.
\newblock In {\em 2017 Seventh international conference on affective computing and intelligent interaction workshops and demos (ACIIW)}, pages 176--180. IEEE, 2017.

\bibitem{xu2018deep}
R.~Xu, Z.~Chen, W.~Zuo, J.~Yan, and L.~Lin.
\newblock Deep cocktail network: Multi-source unsupervised domain adaptation with category shift.
\newblock In {\em Proceedings of the IEEE conference on computer vision and pattern recognition}, pages 3964--3973, 2018.

\bibitem{yan2018unsupervised}
K.~Yan, W.~Zheng, Z.~Cui, Y.~Zong, T.~Zhang, and C.~Tang.
\newblock Unsupervised facial expression recognition using domain adaptation based dictionary learning approach.
\newblock {\em Neurocomputing}, 319:84--91, 2018.

\bibitem{yuan2022self}
J.~Yuan, F.~Hou, Y.~Du, Z.~Shi, X.~Geng, J.~Fan, and Y.~Rui.
\newblock Self-supervised graph neural network for multi-source domain adaptation.
\newblock In {\em Proceedings of the 30th ACM International Conference on Multimedia}, pages 3907--3916, 2022.

\bibitem{zhao2021madan}
S.~Zhao, B.~Li, P.~Xu, X.~Yue, G.~Ding, and K.~Keutzer.
\newblock Madan: multi-source adversarial domain aggregation network for domain adaptation.
\newblock {\em International Journal of Computer Vision}, 129(8):2399--2424, 2021.

\bibitem{zhao2020multi}
S.~Zhao, G.~Wang, S.~Zhang, Y.~Gu, Y.~Li, Z.~Song, P.~Xu, R.~Hu, H.~Chai, and K.~Keutzer.
\newblock Multi-source distilling domain adaptation.
\newblock In {\em Proceedings of the AAAI Conference on Artificial Intelligence}, volume~34, pages 12975--12983, 2020.

\bibitem{zhu2017unpaired}
J.-Y. Zhu, T.~Park, P.~Isola, and A.~A. Efros.
\newblock Unpaired image-to-image translation using cycle-consistent adversarial networks.
\newblock In {\em Proceedings of the IEEE international conference on computer vision}, pages 2223--2232, 2017.

\bibitem{zhu2016discriminative}
R.~Zhu, G.~Sang, and Q.~Zhao.
\newblock Discriminative feature adaptation for cross-domain facial expression recognition.
\newblock In {\em 2016 International Conference on Biometrics (ICB)}, pages 1--7. IEEE, 2016.

\bibitem{zhuang2015supervised}
F.~Zhuang, X.~Cheng, P.~Luo, S.~J. Pan, and Q.~He.
\newblock Supervised representation learning: Transfer learning with deep autoencoders.
\newblock In {\em Twenty-Fourth International Joint Conference on Artificial Intelligence}, 2015.

\end{thebibliography}
}

\end{document}